\documentclass[conference]{IEEEtran}
\IEEEoverridecommandlockouts
\usepackage{cite}
\usepackage{amsmath,amssymb,amsfonts}
\usepackage{graphicx}
\usepackage{textcomp}
\usepackage{xcolor}
\usepackage{microtype}

\def\BibTeX{{\rm B\kern-.05em{\sc i\kern-.025em b}\kern-.08em
    T\kern-.1667em\lower.7ex\hbox{E}\kern-.125emX}}

\begin{document}

\title{Advancing Remote Medical Palpation through Cognition and Emotion%
\thanks{Funded by NEDO grant 21501012-0.}}

\author{
\IEEEauthorblockN{Matti Itkonen\IEEEauthorrefmark{1},
Shotaro Okajima\IEEEauthorrefmark{2},
Sayako Ueda\IEEEauthorrefmark{3},
Alvaro Costa-Garcia\IEEEauthorrefmark{4},
Yang Ningjia\IEEEauthorrefmark{5},\\
Tadatoshi Kurogi\IEEEauthorrefmark{6},
Takeshi Fujiwara\IEEEauthorrefmark{6},
Shigeru Kurimoto\IEEEauthorrefmark{2},
Shintaro Oyama\IEEEauthorrefmark{2},\\
Masaomi Saeki\IEEEauthorrefmark{2},
Michiro Yamamoto\IEEEauthorrefmark{2},
Hidemasa Yoneda\IEEEauthorrefmark{2},
Hitoshi Hirata\IEEEauthorrefmark{2},
Shingo Shimoda\IEEEauthorrefmark{2},~\IEEEmembership{Member,~IEEE}}
\IEEEauthorblockA{\IEEEauthorrefmark{1}School of Computing, University of Eastern Finland, Joensuu, Finland}
\IEEEauthorblockA{\IEEEauthorrefmark{2}Graduate School of Medicine, Nagoya University, Nagoya, Japan}
\IEEEauthorblockA{\IEEEauthorrefmark{3}Dept.\ of Psychology, Japan Women's University, Tokyo, Japan}
\IEEEauthorblockA{\IEEEauthorrefmark{4}Human Augmentation Research Center, AIST, Kashiwa, Japan}
\IEEEauthorblockA{\IEEEauthorrefmark{5}Zhejiang Lab, Hangzhou, China \quad
\IEEEauthorrefmark{6}TOYODA Gosei Co.\ Ltd., Inazawa, Japan}
}

\maketitle

\begin{abstract}
Medical palpation is more than force transmission. It is a bidirectional cognitive and emotional exchange between doctor and patient. We model two complementary touch pathways: active touch by the doctor (kinesthetic and tactile) and passive touch by the patient (subjective and emotional). We use this framework to design a mixed-reality telepalpation prototype and evaluate it with 14 experienced clinicians serving as both doctors and patients across 391 trials. Touch location was transmitted reliably across participants, while 
force perception showed systematic inter-individual variation, 
suggesting that force alone is insufficient to characterize the 
palpation experience.
\end{abstract}


\section{Introduction}
\label{sec:intro}

\IEEEPARstart{M}edical palpation integrates tactile sensation with clinical reasoning, prior experience, and patient responses to assess a patient's condition. While telemedicine has expanded remote healthcare~\cite{haleem_telemedicine_2021}, it does not support physical diagnostic techniques such as palpation. Existing telepalpation systems address this primarily as a force transmission problem, yet palpation involves far more: the doctor's active haptic exploration, the patient's subjective and emotional response, and the diagnostic inference that arises from their interaction.

We introduce the concept of \emph{Palpation Reality Beyond Real}: a remote system designed around the full cognitive loop of palpation rather than its mechanical component alone. Beyond replicating in-person examination, such a system has the potential to exceed it by integrating observations and measurements unavailable to the bare eye and hand. This positions remote palpation within a broader scope than telehaptics, one where cognition and emotion are first-class design targets and where digital mediation becomes a diagnostic advantage rather than a compromise.

\section{Cognitive Framework}
\label{sec:cognition}

\begin{figure}[t]
  \centering
  \includegraphics[width=\linewidth]{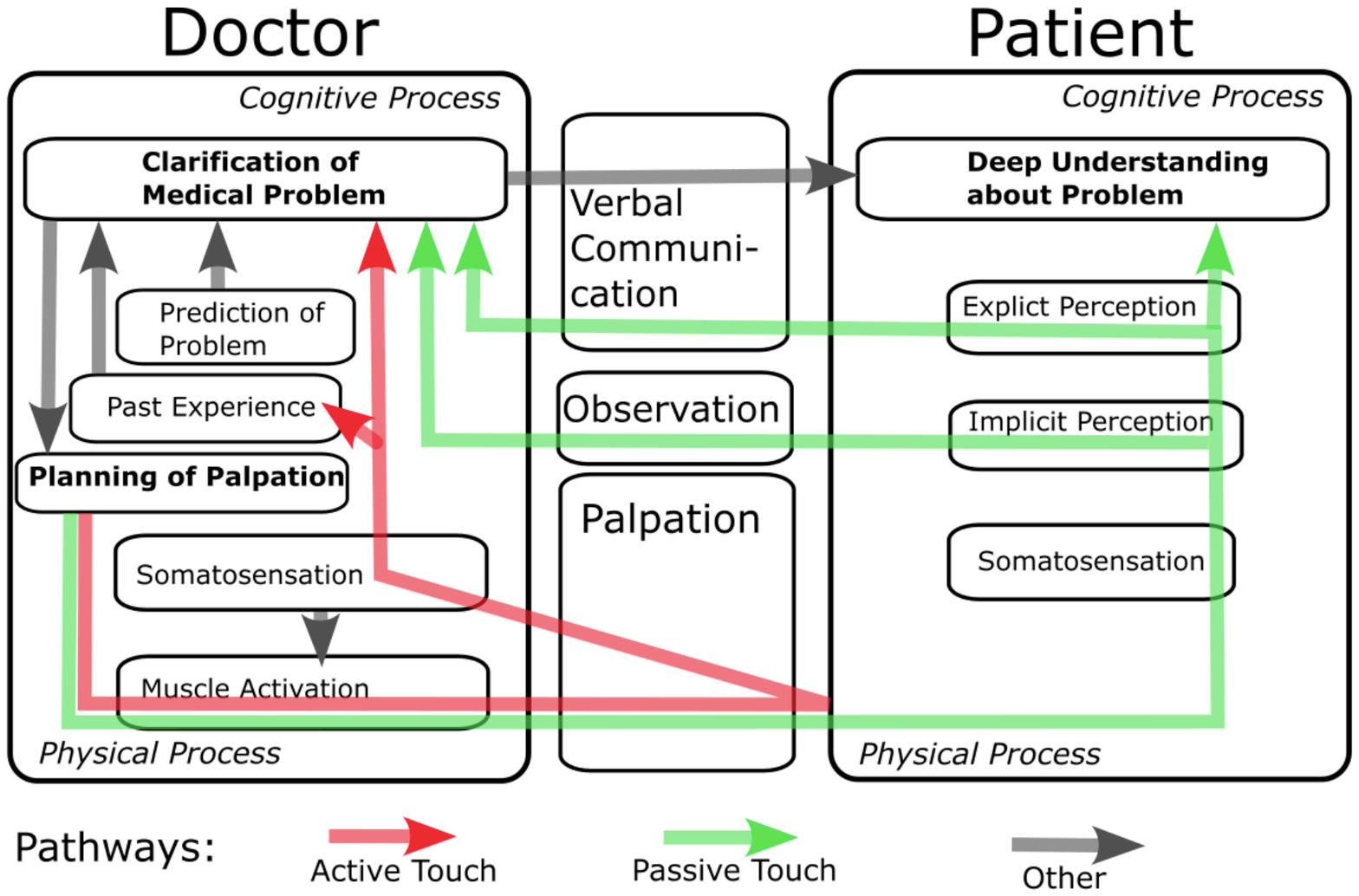}
  \caption{Signal pathways during palpation. Active touch (doctor): tactile and kinesthetic feedback integrates with clinical memory and prior experience to form a diagnostic understanding. Passive touch (patient): the doctor's contact elicits explicit responses (verbally communicated) and implicit responses through facial expression, tone, and posture.}
  \label{fig:pathways}
\end{figure}

\textbf{Active touch} (doctor side) involves stereotypical exploratory movements~\cite{lederman_extracting_1993} coupled with proprioceptive and kinesthetic feedback. Tactile sensation links current perception to clinical memory, supporting abductive diagnostic reasoning~\cite{barrows_clinical_1987}. Haptic memory encodes experience durably~\cite{hutmacher_longterm_2018}, making touch a powerful anchor for clinical cognition.

\textbf{Passive touch} (patient side) captures subjective sensations elicited by the doctor's contact~\cite{passivetouch}. Explicit responses such as pain are verbally reported; implicit ones emerge through facial expression, tone, and posture. Affective touch, mediated by C-tactile afferents~\cite{mcglone_discriminative_2014}, underlies the emotional dimension of patient response.

\section{System}
\label{sec:implementation}

\begin{figure}[t]
  \centering
  \includegraphics[width=\linewidth]{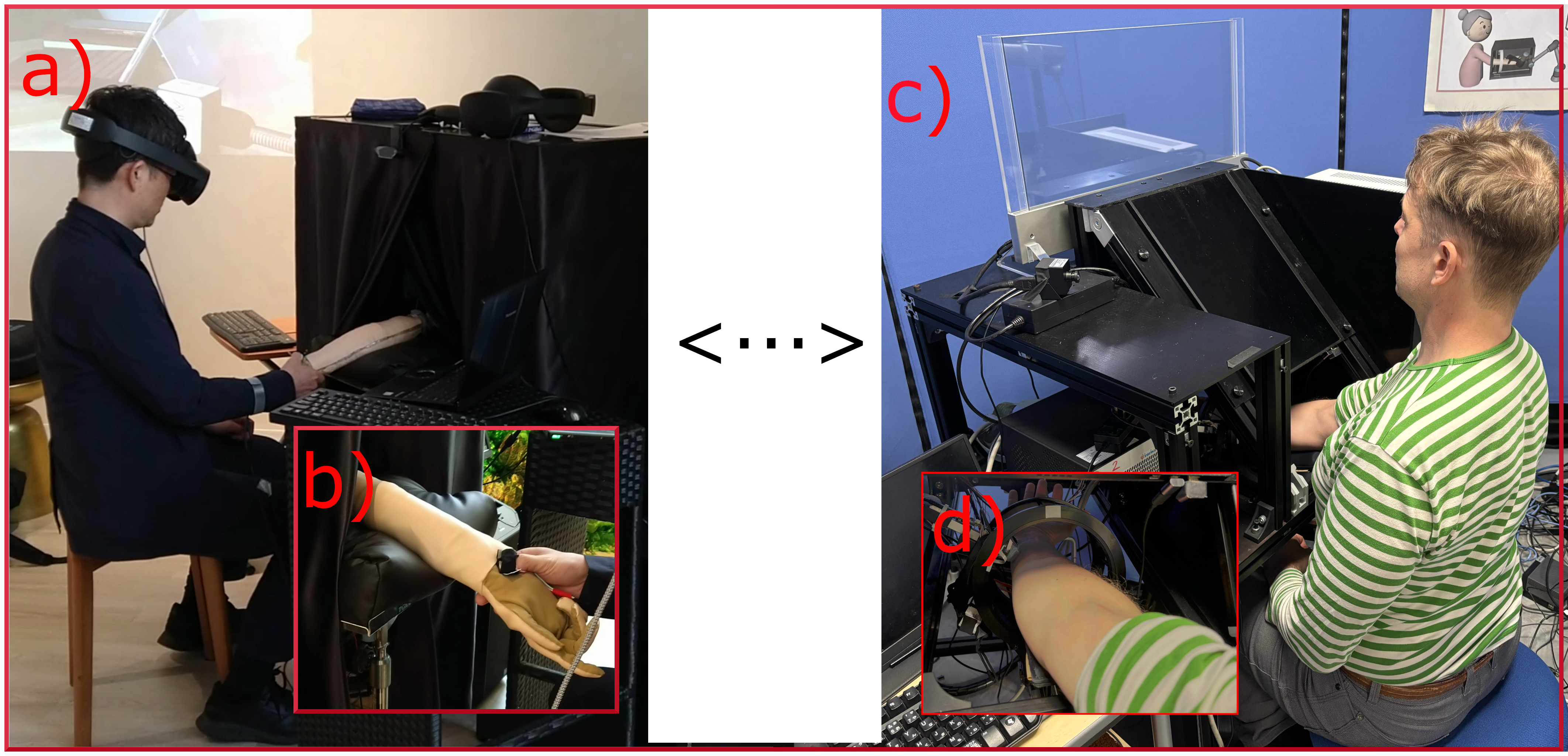}
  \caption{Prototype setup. \textbf{a)} Doctor in VR headset interacts with patient replica via tactile finger displays. \textbf{b)} Patient replica. \textbf{c)} Patient environment: dual displays show the doctor's face and augmented hands overlaid on the patient's own hand. \textbf{d)} Rotating gantry with linear probing actuator.}
  \label{fig:system}
\end{figure}

The prototype (Fig.~\ref{fig:system}) implements both pathways over a telecommunication link. On the doctor's side, a physical patient replica acts as a haptic controller; tactile displays on the fingers~\cite{kurogi_haptic_2019} preserve natural palpation feel, leveraging existing haptic memory without device-specific training. A head-mounted display provides a mixed-reality view of the patient.

On the patient's side, a robotic manipulator reproduces the doctor's actions in real time. Dual displays show the doctor's face and the patient's hand augmented with the doctor's hands---supporting passive touch and nonverbal communication. Live facial video is streamed bidirectionally throughout.

A key design principle is that the doctor should not need to learn new interaction paradigms. Unlike typical teleoperation systems that impose their own haptic vocabulary, the replica-based approach aligns with natural palpation kinematics so that existing clinical haptic memory transfers directly. Latency compensation and anticipatory motion control~\cite{Okajima2} are handled at the system level, keeping the cognitive load on the doctor focused on diagnosis rather than device operation.

\section{Feasibility Evaluation}
\label{sec:performance}

\subsection{Method}
Fourteen experienced clinicians formed 44 doctor--patient pairs and conducted 391 palpation trials. The doctor palpated three wrist landmarks: Carpal Tunnel (CT, ganglion cyst), Radial Artery (RA, hemangioma), and Radial Styloid Process (RSP, De Quervain syndrome), at three force levels reflecting each doctor's own clinical judgment of weak, medium, and strong palpation. The participant acting as patient rated perceived spatial location and force on a 10-point Likert scale. The sequence of cases and pressures was consistent across trials. 

\subsection{Results and Discussion}

\begin{table}[t]
\centering
\caption{Descriptive statistics for force and position scores.}
\label{tab:scores}
\renewcommand{\arraystretch}{1.1}
\begin{tabular}{llrrrrr}
\hline
Target & Type & $n$ & \multicolumn{2}{c}{Force} & \multicolumn{2}{c}{Position}\\
       &      &     & Mean & SD & Mean & SD \\
\hline
Weak   & Force    & 129 & 8.72 & 1.30 & 8.16 & 1.77 \\
Mid    & Force    & 130 & 7.71 & 1.62 & 8.12 & 1.74 \\
Strong & Force    & 132 & 7.60 & 2.11 & 8.22 & 1.68 \\
CT     & Position & 130 & 7.98 & 1.88 & 8.23 & 1.62 \\
RA     & Position & 132 & 8.08 & 1.69 & 7.98 & 1.86 \\
RSP    & Position & 129 & 7.97 & 1.77 & 8.29 & 1.69 \\
\hline
Total  &          & 391 & 8.01 & 1.78 & 8.16 & 1.73 \\
\hline
\end{tabular}
\end{table}

Position scores were uniformly high (mean~8.16) with no significant 
differences across landmarks (all $p>0.41$), confirming that the 
system conveys touch location with sufficient fidelity for clinical 
orientation, a prerequisite for any meaningful palpation encounter. Force level had a significant effect on force ratings ($F(2,385)=4.53$, $p=0.011$), though pairwise comparisons were not significant after Tukey adjustment (all $p>0.06$). The absence of pairwise differences likely reflects inter-individual variation in how force was applied and perceived: doctors set force levels according to their own clinical judgment, and patients rated them through their own sensory experience.

\section{Conclusions}
\label{sec:conclusion}

Standardizing force alone cannot resolve the variability observed in palpation perception; the full cognitive and emotional context in which touch is given and received shapes the experience on both sides. The pathways framework and prototype presented here offer a broader foundation for this challenge than conventional telehaptics approaches.

The current evaluation is an intentional first step that tests whether the channel works before asking what can be sent through it. What remains unaddressed is substantial: ratings were subjective, no baseline comparison against in-person or force-only telepalpation was made, and the patient-side emotional and trust responses central to the passive touch pathway were not measured. These are not incidental gaps but the core of what the framework motivates.

The research agenda is therefore clear: does transmitting cognitive and emotional signals alongside haptic force change what doctors understand and what patients can express? The augmented reality and social touch~\cite{gallace_science_2010} dimensions of the system offer particular promise for enriching the patient-side experience in ways that purely force-based approaches cannot.

\bibliographystyle{IEEEtran}
\bibliography{cbms_palpation}

\end{document}